\documentclass[acmtog]{acmart}
\AtBeginDocument{%
  \providecommand\BibTeX{{%
    \normalfont B\kern-0.5em{\scshape i\kern-0.25em b}\kern-0.8em\TeX}}}

\acmSubmissionID{599}

\setcopyright{acmlicensed}
\acmJournal{TOG}
\acmYear{2024} \acmVolume{43} \acmNumber{4} \acmArticle{117} \acmMonth{7}\acmDOI{10.1145/3658184}

\setcitestyle{nosort} 

\usepackage{amsmath}
\usepackage{tabularx}
\usepackage{colortbl}
\usepackage{algorithm} 
\usepackage{algpseudocode} 
\usepackage{algorithmicx}
\usepackage{wrapfig}
\usepackage{adjustbox}
\usepackage{wrapfig}
\usepackage{soul}
\usepackage{longtable}
\usepackage{mathtools}
\usepackage{tabu}
\mathtoolsset{centercolon} 
\usepackage{booktabs}
\usepackage{colortbl}
\usepackage{siunitx}

\newcommand{\defX}{\mathbf{x}}
\newcommand{\refX}{\mathbf{X}}
\newcommand{\dofs}{\mathbf{z}}

\newcommand{\R}{\mathbb{R}}
\newcommand{\W}{\mathbf{W}}
\newcommand{\dofsMat}{\mathbf{Z}}

\newcommand{\dofsMatN}{\mathbf{Z}}
\newcommand{\WOpt}{\W_{\Params^*}}

\newcommand{\NDist}{\mathcal{N}}
\newcommand{\Occupancy}{\Phi}
\newcommand{\mass}{\mathbf{M}}
\newcommand{\NumHandle}{n}
\newcommand{\Jacobian}{\mathbf{J}}
\newcommand{\Params}{\theta}
\newcommand{\DeformationMap}{\phi}
\newcommand{\MaterialEnergy}{\Psi}
\newcommand{\ElasticPotential}{E_\textrm{pot}}
\newcommand{\LElastic}{\mathcal{L}_\textrm{elastic}}
\newcommand{\LOrtho}{\mathcal{L}_\textrm{ortho}}
\newcommand{\WElastic}{\lambda_\textrm{elastic}}
\newcommand{\WOrtho}{\lambda_\textrm{ortho}}

\newcommand{\ourmethod}{{Simplicits}}

\newcommand{\dl}[1]{\textcolor{black}{#1}}
\newcommand{\rev}[1]{\textcolor{black}{#1}}

\newcommand{\eg}{\emph{e.g.}} 
\newcommand{\ie}{\emph{i.e.}} 

\renewcommand{\eqref}[1]{Equation~\ref{eq:#1}}

\newcommand{\secref}[1]{Section~\ref{sec:#1}}

\begin{document}

\title{Simplicits: Mesh-Free, Geometry-Agnostic, Elastic Simulation}


\author{Vismay Modi}
\affiliation{%
  \institution{University of Toronto}
  \city{Toronto}
  \country{Canada}
}
\email{vismay@cs.toronto.edu}

\author{Nicholas Sharp}
\affiliation{%
  \institution{Nvidia}
  \city{Seattle}
  \country{USA}
}
\email{nmwsharp@gmail.com}

\author{Or Perel}
\affiliation{%
  \institution{Nvidia}
  \city{Tel Aviv}
  \country{Israel}
}
\email{operel@nvidia.com}

\author{Shinjiro Sueda}
\affiliation{%
  \institution{Texas A\&M University}
  \city{College Station}
  \country{USA}
}
\email{sueda@tamu.edu}

\author{David I. W. Levin}
\affiliation{%
  \institution{University of Toronto}
  \city{Toronto}
  \country{Canada}
}
\email{diwlevin@cs.toronto.edu}

\renewcommand{\shortauthors}{Modi et al.}

\begin{abstract}    
The proliferation of 3D representations, from explicit meshes to implicit neural fields and more, motivates the need for simulators agnostic to representation. We present a data-, mesh-, and grid-free solution for elastic simulation for any object in any geometric representation undergoing large, nonlinear deformations. We note that every standard geometric representation can be reduced to an occupancy function queried at any point in space, and we define a simulator atop this common interface. For each object, we fit a small implicit neural network encoding spatially varying weights that act as a reduced deformation basis. These weights are trained to learn physically significant motions in the object via random perturbations. Our loss ensures we find a weight-space basis that best minimizes deformation energy by stochastically evaluating elastic energies through Monte Carlo sampling of the deformation volume.
At runtime, we simulate in the reduced basis and sample the deformations back to the original domain.
Our experiments demonstrate the versatility, accuracy, and speed of this approach on data including signed distance functions, point clouds, neural primitives, tomography scans, radiance fields, Gaussian splats, surface meshes, and volume meshes, as well as showing a variety of material energies, contact models, and time integration schemes.
\end{abstract}


\begin{teaserfigure}
  \includegraphics[width=\textwidth]{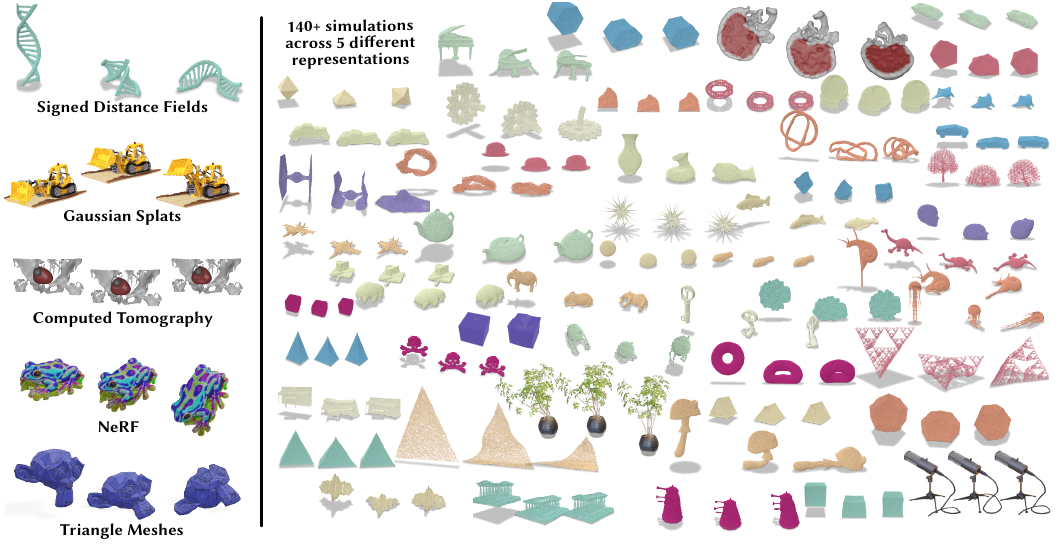}
  \vspace*{-1em}
  \caption{Mesh-free volumetric simulation of objects represented by explicit triangle meshes, point clouds, implicit functions, Computed Tomography volumes, NeRFs, and Gaussian Splats, all produced using our data-free neural fields-based simulation algorithm. Here we show frames from 60 of the 140+ simulations performed.}
  \Description{Description.}
  \label{fig:Teaser}
\end{teaserfigure}

\maketitle 
\section{Introduction}
Across visual computing, there is ever-growing use of an incredible variety of 3D representations, from explicit meshes to implicit neural fields, each a source of important and high-fidelity geometric content. Entire databases of implicit shapes, neural or otherwise, are readily available and contain objects ranging from simple polyhedra to wildly complex fractals. \rev{One of the great advancements in computer graphics is the ease with which any user, expert or novice, can convincingly render any and all of these geometric representations to display rich 3D scenes.} This project (\ourmethod{}) is an attempt to bring the benefits of representation agnostics to elastic simulation.

Functional, robust, feature-rich physics-based elastic simulators are often closely connected to one input geometry type. Once we start to consider other inputs, that bespoke toolchain must change, in potentially complicated ways. Even particle-based methods such as material point method (MPM) and smoothed-particle hydrodynamics (SPH), which reduce surface-to-volume conversion to point sampling, often struggle to resolve intricate boundaries and can exhibit artifacts in the simulated motion.

\ourmethod{} alleviates these issues and provides \emph{mesh-free, reduced, physics-based elastic simulation}. \ourmethod{} is built upon the observation that any geometric representation is encoded either explicitly or implicitly by an inside-outside (occupancy) function. \rev{For most representations, occupancy functions are either trivial or the subject of well-studied algorithms: the sign of the signed-distance function of a mesh, likewise querying an SDF field, fast winding numbers on a point cloud (\cite{Barill2018Winding}), etc. For NeRFs/Splats and medical scans we extract occupancy from the density field.}

Our algorithm is mesh-free at every stage, relying on linear blend skinning (LBS) to characterize the shape-aware/boundary-aware deformation of our object. \rev{Skinning weights are represented via a neural field over the object (in $\R^3$) and are optimized during training after random initialization. Our approach deviates from standard LBS since we do not explicitly define handle locations on the object. Our skinning weights need not satisfy partition-of-unity or the Kronecker delta properties. Instead, handles are simply transformation matrices applied over the object scaled by skinning weights as described by~\citet{benchekroun2023fast}}.

\begin{figure}[b]
     \centering
     \includegraphics[width=\columnwidth]{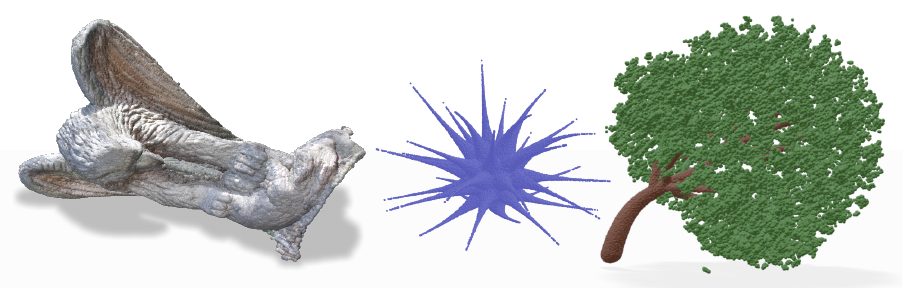}
     \caption{Simulations of point clouds undergoing large deformations. Our method produces shape-aware skinning weights on complex geometries.
     \label{fig:PC_Dino_Spike_Tree}
     }
\end{figure}

Our method can capture the nuanced behavior of geometrically complex shapes (\autoref{fig:PC_Dino_Spike_Tree}) as well as heterogeneous materials (\autoref{fig:CT_Brain}) by using “skinning weights” as a physics-informed deformation basis. Handle transformation DOFs robustly capture large rotations and deformations without artifacts. At runtime, we use Newton's method to solve for handle matrix values that give an optimal result according to the integration equations.

To demonstrate the efficacy of \ourmethod{}, we perform a large number of elastodynamic simulations on a myriad of input geometry representations---triangle meshes, signed and unsigned distance functions, neural implicits, NeRFs Gaussian Splats, and medical imagery---all without modifying the algorithm or its parameters. Across these representations, we show results that are volumetric, thin rods or sheets, and shapes that combine all three of these challenging features---again without requiring the algorithm to make any explicit distinction between them. In total this paper features more than 140 simulation results. It is our hope that the generality of this approach helps to make elastodynamic physics simulation a more accessible and enjoyable tool for the greater graphics community, rather than just the expert practitioners.

\begin{figure*}
    \vspace{1em}
     \centering
        \includegraphics[width=1\textwidth]{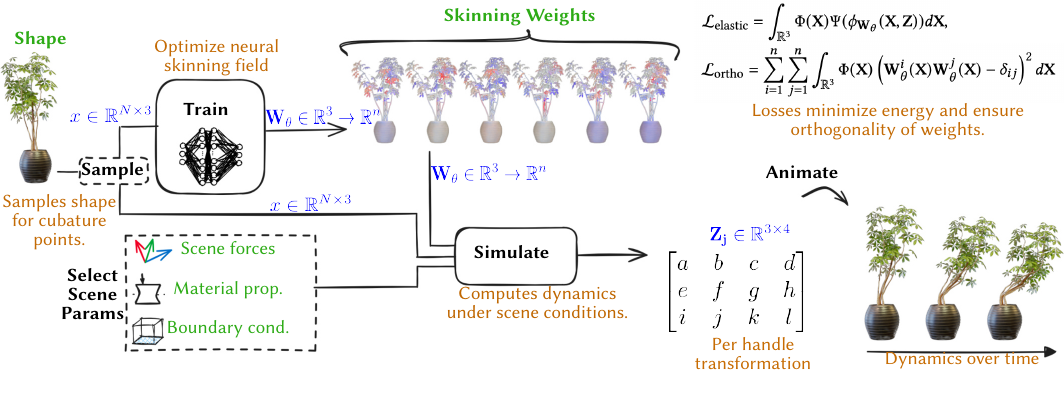}
     \caption{Pipeline overview. First, the skinning weights $\W_\Params$ are learned by minimizing the potential and orthogonality losses over randomized deformations ($N$ is the number of samples, and $n$ is the number of skinning handles). Then, given physical material properties and scene conditions, keyframes of transformation-per-handle $\dofsMat_j$ are generated, finally combined into an animated object.}
     \label{fig:overview}
\end{figure*}

\section{Related Work}
\label{sec:related_work}

\paragraph{Mesh-Based Methods}
Mesh-based elasticity simulation has long been a part of computer graphics~\citep{Terzopoulos1987}. The most common variant in use today, for volumetric simulation, is the linear tetrahedral finite element method~\citep{cutler2002procedural}. Since its initial introduction to graphics there have been fundamental improvements in performance~\citep{bargteil2007finite,bouaziz2014projective,macklin2016xpbd}, extensions to more complicated phenomena~\citep{bargteil2007finite} including contact~\citep{Li2020IPC} and higher-order implementations~\citep{Schneider:2019:PFM,Schneider:2018:DSA}. In our context, the most important advancements have been made in the geometry processing pipeline that constructs volumetric tetrahedral meshes from surface input. Here robust tools~\citep{Hu2018,diazzi2023constrained} enable virtually closed-box simulation of such geometry. 
However, even broadening the scope of input geometry slightly increases the complexity of the geometry processing and simulation pipelines significantly. For instance, objects with thin features either force volumetric mesh resolutions to be exceedingly high (impractical for many applications) or require special co-dimensional treatments that allow for the inclusion of medial axis geometry (lines for rods and sheets for triangles) into the simulation. Specialized simulators exist for geometry solely constructed from thin primitives~\citep{baraff2023large,bergou2008discrete} but, coupling these to volumetric approaches requires non-trivial solutions~\citep{chang2019unified,Li2021CIPC}.

\paragraph{Embedded Methods}
An attractive alternative is embedded methods, which wrap geometry inside of some easily constructable discretization for simulation~\citep{joshi2007harmonic,longva2020higher,lee2018skinned}, and have even been extended to recent neural representations~\citep{yuan2022nerf,xu2022deforming,garbin2022voltemorph}.
These can produce excellent results, but will naively couple disparate geometric elements embedded in the same element, again necessitating additional algorithmic modifications~\citep{nesme2009preserving}.

\paragraph{Mesh-Free Methods}
Alternatively, to simulate complex geometries, one could turn to mesh-free methods, which were first introduced to graphics by \citet{Desbrun1995}. 
A well-known example is smoothed-particle hydrodynamics (SPH) \cite{Desbrun1996,peer2018implicit, kugelstadt2021fast}, which requires the connectivity to be updated at every time step.
\ourmethod{}, on the other hand, does not need neighborhood information to be queried at runtime. \rev{Instead, the deformation gradient and other derived quantities, such as strain and stress, are stored implicitly within a neural field and can be computed anywhere inside the object.}
The material point method (MPM) has also become popular in graphics in the past decade \cite{Jiang2016mpm,wolper2019cdmpm,hu2019taichi}.
However, it still requires an explicit background mesh/grid and is prone to numerical artifacts that prevent accurate (quantitatively or qualitatively) simulation of large-scale elastodynamics of complex geometries. \rev{Concurrent work for direct simulation of Gaussian Splats~\citep{xie2023physgaussian} inherits these limitations.} 
Moving least squares (MLS) formulations were introduced in graphics by \citet{Muller2004point} and later extended by \citet{martin2010unified} to enable a unified simulation framework for volumetric and co-dimensional objects.
However, these methods require extremely dense sampling, which can hurt performance. Also, their interpolation functions do not take shape-aware distance into account and so can couple close but disconnected parts of an object.
\citet{faure2011sparse} introduced a frame-based mesh-free approach, along with pre-computed weight functions to represent object kinematics. This approach is closest to ours technically and philosophically; however, it still requires an explicit boundary for boundary conditions which means it cannot directly take, as input, representations such as Gaussian Splats, while our proposed approach can. Furthermore, the method requires a dense voxelization of the object for computing and storing weight functions which is memory intensive. \rev{Another concurrent approach by \citet{feng2023pienerf} develops a voronoi-cell based method for discretizing and simulating NeRFs, however the cell discretizations heavily affect the dynamics and shape-awareness of the simulations.} Many other methods have been developed to represent dynamic NeRFs, often using deformation fields~\cite{park2021nerfies,pumarola2021d}, but these are designed to fit supervised observations, rather than to simulate new physical behavior.

\paragraph{Neural Fields for Physics}
Leveraging neural networks and neural fields for physics simulation is a fast-growing research area. \citet{chenwu2023insr-pde} develop a neural simulation technique that can handle infinitely high geometric resolution at a very high simulation cost per time step (often on the order of hours). Other neural techniques pioneered by \citet{Fulton2019} can alleviate some of the issues related to simulation speed via reductions to a nonlinear, neural latent space. Recently, this idea was greatly extended to Continuous-Reduced Order Models (CROM)~\cite{chen2023crom,chang2023licrom} which use prior simulation data to generate reduced order models for simulation. \rev{\dl{While the method by \citet{chang2023licrom} demonstrates cutting,} it relies on the availability of pre-simulated datasets, thus inheriting the geometry processing challenges described above. Additionally, LiCROM requires several hours of training (while ours requires a couple of minutes) and CROM occasionally requires explicit mesh gradients. This again limits their applicability without further algorithmic development. In contrast, our data-free, self-supervised training over a physics-loss is more similar to~\cite{zehnder2021ntopo, sharp2023data}. More specifically, \citet{zehnder2021ntopo} focus on topology optimization over a continuous neural field, while \citet{sharp2023data} optimize a neural network to learn the low-order subspace of an explicit mesh for \textit{kinematics}. Unfortunately, \dl{while this method succeeds in generating kinematic subspaces, it struggles with convergence during dynamics.}} In contrast, we use data-free, nonlinear skinning weight optimization to learn skinning weight functions over complex objects whether they are represented by an explicit mesh or not.

\paragraph{Skinning Modes}
\rev{Neural skinning fields using a multi-layer perceptron were originally proposed by \citet{saito2021scanimate} for animated characters and subsequently extended by \citet{mihajlovic2021leap} to learn occupancies of articulated characters for collision resolution.} Our work on neural skinning fields for dynamics is most related to~\cite{benchekroun2023fast,trusty2023subspace}, a method for skinning-based dynamics simulation. In particular, as a starting point, the skinning eigenmodes by \citet{benchekroun2023fast} and extend them in three ways. First, we extend the eigenmode optimization into the nonlinear regime, allowing for more complex energies. Second, in this nonlinear regime, we show that it is possible and preferable to optimize the weights under the assumption of full affine transformations, rather than just the translations of prior work. Next, rather than store skinning weights on a volumetric tetrahedral mesh, we store the weights as neural fields enabling a fully meshless pipeline from beginning to end. Finally, we use this weight field to simulate stable elasto-dynamics using the implicit Euler time integrator~\cite{Gast2015, martin2011example}.

\section{Method}

Our goal is to construct an integrator for simulating time-varying elastodynamics. Our method will take as input a rest-state object, defined in any geometric representation that supports evaluating an ``inside-outside'' density $\Occupancy(x) \in \R^3 \rightarrow \R$ such that $\Occupancy(x) = 1$ inside the object and $\Occupancy(x) = 0$ outside, possibly with a blurry boundary in-between. Our output will be a set of neural fields that encode skinning weights suitable for dynamics simulation (\autoref{fig:overview}).

\subsection{Implicit Time Integration}
We start with the deformation map  $\defX = \DeformationMap(\refX, \dofs(t))$, where
$\refX \in \R^3$ is a point in reference space and $\defX\in\R^3$ is its deformed position according to $\dofs(t)$, a time-varying vector of as-yet unspecified degrees-of-freedom (DOFs). If $\DeformationMap$ is a linear function with respect to $\dofs$, then we can discretize standard implicit time integration as the following optimization problem
\begin{equation}
    \dofs_{t+1} = \underset{\dofs}{\arg\min} \frac{1}{2}\|\dofs - \tilde{\dofs_{t}}\|^2_\mass + h^2\ElasticPotential(\dofs),
    \label{eq:time_stepping}
\end{equation} where $\dofs_{t+1}$ are the integrated DOFs for the next time step, $\|\cdot\|^2_\mass$ is the squared norm weighted by an appropriate mass matrix, $h$ is the simulation time step, $\ElasticPotential$ is the elastic potential energy of the simulated object and $\tilde{\dofs}_{t}$ is the standard first order predictor for $\dofs$. This formulation can be augmented with constraints and penalty springs for maintaining fixed or moving boundary conditions along with barrier functions to handle contact and then solved using standard Newton-based methods ~\cite{NoceWrig06, Li2020IPC}. Our goal is to choose an appropriate set of DOFs that are both expressive enough to generate compelling simulation results and amenable to a neural representation for adaptive generality.

\subsection{Degrees-of-Freedom}
As previously noted, simply parameterizing $\DeformationMap$ as a pure neural nonlinear function typically fails for dynamics~\citep{sharp2023data}. 
Standard FEM shape functions satisfy linearity, but require explicit polyhedra for construction. 
Particles with extrapolating shape functions are more general, but incur modeling and numerical challenges as discussed in \secref{related_work}.
Blended affine transformations, or \emph{skinning handles}, are promising solution, offering easy numerical integration and accurate modeling of elastic phenomena.
Accordingly, we parameterize our deformation map, $\phi$ using linear blend skinning
\begin{equation} \label{eq:lbs_equation}
   \DeformationMap(\refX, \dofs) = \refX + \sum_{j=1}^\NumHandle \W_{\!j} (\refX) \dofsMat_j  \begin{bmatrix} \refX \\ 1 \end{bmatrix}
\end{equation}
where $\NumHandle$ is the number of skinning handles, $\W_{\!j}(\refX) \in \R^3 \rightarrow \R$ is the spatially varying scalar shape function associated with handle $j$, $\dofsMat_j\in\R^{3\times4}$ is the $j^{th}$ skinning handle. The vector $\dofs$ of DOFs is formed by flattening the stacked handle transforms: $\dofs=\text{flat}(\dofsMat) \in \R^{12n}$. In the subsequent text, we assume that $\dofs \in \R^{12n}$ and $\dofsMat \in \R^{(3 \times 4) \times n}$ are interchangeable, with implicit conversion between them.

\setlength{\columnsep}{1em}
\setlength{\intextsep}{0em}
\begin{wrapfigure}{r}{110pt}
\includegraphics[width=110pt]{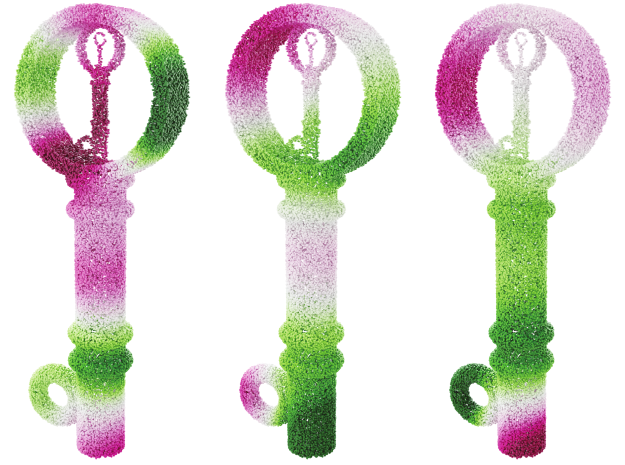}
    \caption{
        Trained skinning weights.\label{fig:key_modes}
    }
\end{wrapfigure}
While previous work stores the skinning shape functions $\W \in \R^3 \rightarrow \R^n$ on an explicit volumetric mesh~\cite{benchekroun2023fast}, we instead store these functions as a vector-valued continuous neural field $\W_\Params \in \R^3 \rightarrow \R^n$, removing the need for any explicit geometric scaffolding, yielding a purely meshless discretization during training and at runtime. 

\subsection{Meshless Integration in Space}
\label{sec:meshless_integration}
Before we address the training of these neural fields, we must first describe the evaluation of integral quantities over the domain.
Both our mass matrix and potential energies are quantities integrated over the domain of the geometry
\begin{equation}
\mass = \int_\Omega \rho\Jacobian(\refX)^T\Jacobian(\refX) d\Omega
\qquad
\ElasticPotential = \int_\Omega \MaterialEnergy(\DeformationMap(\refX)) d\Omega,
\label{eq:mass_matrix_and_potential_energy}
\end{equation}
where $\Omega$ is the interior domain of our object in the undeformed reference space, $\rho\in\R$ is its physical density, $\Jacobian = \nabla_{\!\dofs}\DeformationMap$ is the Jacobian function of the deformation map evaluated at $\refX \in \Omega$, and $\MaterialEnergy$ is the strain energy density function \citep{kim2020dynamic}. We note that since the deformation map $\DeformationMap$ is linear with respect to the DOFs $\dofs$, the Jacobian $\Jacobian \in \R^{3 \times 12n}$ and the mass matrix $\mass \in \R^{12n \times 12n}$ are constant.

Inserting our occupancy density into any such integral yields the general from 
\begin{equation}
G = \int_{\R^3} \Occupancy(\refX) g(\refX) d\refX,
\end{equation}
where $g$ is the quantity to be integrated, as in Equation \ref{eq:mass_matrix_and_potential_energy}.
We evaluate this via Monte Carlo integration, sampling the domain of the object. 

\subsection{Neural Skinning Field Loss}
Good skinning weights accurately capture large, physically plausible rotations and deformations of an object, even with a limited number of handles.
We seek to fit an implicit weight function $\W$ with these properties; \ie{} we train the parameters $\Params$ of a neural network $\W_\Params$:
\begin{equation}
  \label{eq:objective_combined}
  \Params^* = \underset{\Params}{\arg\min} \ \WElastic \LElastic + \WOrtho \LOrtho.
\end{equation}
The result is a fitted implicit weight function $\WOpt \in \R^3 \rightarrow \R^n$ for the shape, which smoothly captures large nonlinear deformations.

\paragraph{Physical-Plausibility}
Material-aware and spatial-aware deformations have low elastic potential energy (\eqref{mass_matrix_and_potential_energy}), while undesirable nonsensical deformations have very high elastic energy.
Accordingly, we seek weights that minimize the elastic energy for any random set of handle transformations
\begin{equation}
  \label{eq:objective_elastic}
  \LElastic = \int_{\R^3} \Occupancy(\refX) \MaterialEnergy(\DeformationMap_{\W_\Params}(\refX,\dofsMatN)) d\refX,
\end{equation}
where in-practice small transformations $\dofsMatN$ are randomly sampled from $\dofsMatN$ during training, as described later in \autoref{sec:implementation}.

\paragraph{Orthogonality} 
Unfortunately, naively minimizing $\LElastic$ immediately collapses to a trivial solution, with all handles encoding a constant weight, limiting the model to rigid motions.
This could be mitigated in a ``supervised'' manner, by requiring the skinning weights to reproduce some dataset of motions, but such datasets are rarely available in practice.
Instead, we adopt a data-free approach that additionally seeks weights that are mutually orthonormal under the inner product on scalar functions $\int\! f(x) g(x) dx$, which amounts to minimizing
\begin{equation}
  \label{eq:objective_orthogonal}
  \rev{
  \LOrtho = \sum_{i=1}^\NumHandle \sum_{j=1}^\NumHandle\int_{\R^3} \Occupancy(\refX)  \left(  \W_\Params^i(\refX)\W_\Params^j(\refX) - \delta_{ij} \right)^2 d\refX,
  }
\end{equation}
where $i$ and $j$ are handle indices, and $\delta_{ij}$ is the Kronecker delta.



\section{Implementation}
\label{sec:implementation}
\subsection{Network Architecture and Training}
We represent the skinning weight function $\W$ as a small neural network per-object $\W_\Params : \R^3 \to \R^\NumHandle$ with parameters $\Params$, which takes spatial coordinates as input and outputs skinning weights at the point.
Please see the supplemental material for a detailed listing of all parameters.

\paragraph{Architecture}
We choose $\W_\Params$ to be a small multi-layer perceptron (MLP), with ELU activation functions on hidden layers.
The size of the network depends on the complexity of the problem; 9 hidden layers is typical, we use width 64 for all objects.
For the sake of parameter scaling, all inputs are normalized to have length-scale $\approx\!1$ before network evaluation.
Unlike classic skinning formulations, where the weights $\W$ are defined as a partition of unity, or at least positive, we find it effective in practice to leave $\W$ unconstrained in $\R^\NumHandle$, and use them as a reduced subspace, akin to skinning eigenmodes~\cite{benchekroun2023fast}. 

\paragraph{Training}
We fit $\W_\Params$ for each object by minimizing the elastic and orthogonality losses (\eqref{objective_combined}) using the Adam optimizer~\cite{kingma2014adam} with a linearly scheduled learning rate.
For each training iteration, we sample a batch of random transformations $\dofsMatN$ for each handle, as well as cubature points to evaluate spatial integrals (\secref{meshless_integration}).
The transformation matrices $\dofsMatN$ are drawn elementwise from $\dofsMatN\!\sim\!\NDist(\mu, 0)$. We find that $\mu = 0.1$ is effective for unit-scaled objects, and that training on relatively small unconstrained perturbations yields weights which generalize well to large nonlinear deformations.
We schedule the elastic energy as $\MaterialEnergy = (1 - \alpha) \MaterialEnergy_\textrm{linear elastic} + \alpha \MaterialEnergy_\textrm{neohookean}$, where $\alpha$ goes $0 \to 1$ during training, because linear elasticity has stable gradients even for randomly initialized weights, while Neohookean elasticity improves large-deformation behavior of the final weights.

\begin{figure}[b]
  \centering
  \includegraphics{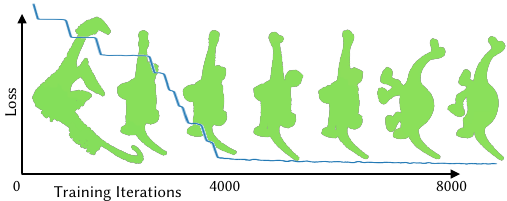}
  \caption{Simulation quality improves during neural skinning weight training.}
  \label{fig:trainingIterationConvergence}
\end{figure} 

\begin{figure*}
    \centering
    \includegraphics{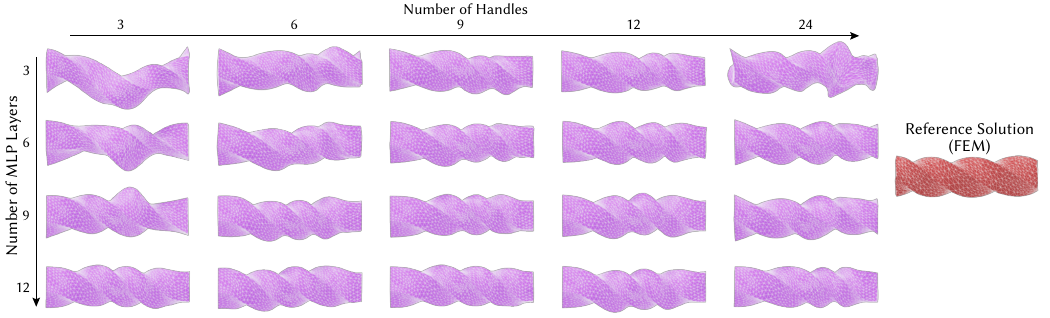}
    \vspace*{-1em}
    \caption{
      Increasing the degrees of freedom in our simulation (number of skinning handles $\NumHandle$) and the capacity of the neural skinning function (number of MLP hidden layers) both increase the expressivity of the method. \rev{Note that network capacity must be sufficient to match the DOFs. Having 3 layers with 24 handles is insufficient for complex deformations. We use nonlinear Neohookean energy for these deformations.}
      \label{fig:ablation}
    }
\end{figure*}



\subsection{Time-Stepping}
Once the skinning weights $\W_\Params$ have been fit, we timestep the simulation by solving \autoref{eq:time_stepping} using standard projected Newton's method with a line search.
For fast simulation we evaluate integrals at a fixed set of sample points drawn once in the rest space as a preprocess.
We emphasize that with this setup, our method does not require evaluating the neural network at all within the physics time-stepping loop, just once as a preprocess to compute weights and their derivatives at sample points---this is a key reason for the speed and robustness of our method.
If desired, the neural skinning weights can later be evaluated at any point in space to query the smooth simulated deformation.
Our approach is compatible with any hyperelastic material, and the energy used for simulation need not match the energy used for training.
The majority of our examples use the stable Neohookean model~\citep{kim2020dynamic}.



\subsection{Representation-Specific Concerns}

A primary goal of this work is to simulate on a very wide variety of representations our experiments include meshes, NeRFs, CT scans, Gaussian Splats, and more.
An in-depth introduction of each of these representations is beyond the scope of this document, but our general procedure is the same for all representations.
We define an occupancy function $\Occupancy$ such as an inside-outside test on a mesh, thresholded NeRF density, or clipped Houndsfield units in a CT scan, as well as specifying a bounding domain for the rest geometry, and any needed physical parameters like density and stiffness.
Spatial samples to evaluate integrals can come directly from the geometry, such as fast mesh sampling and particle subsets from particle-based methods, or be generated by rejection sampling, taking uniformly random points within the domain where $\Occupancy(\refX) > 0$.

After simulation, one generally needs to render or otherwise interact with the deformed object.
For explicit representations such as meshes and particles, our continuous forward deformation field (\eqref{lbs_equation}) can be queried at any point to translate rest points to the deformed location.
If needed, we can also read off local rotations, \eg{} to rotate the orientation of Gaussian Splat particles.
Implicit representations such as NeRFs are more difficult, as volumetric rendering generally requires the reverse map taking points in deformed space back to the rest space; prior work has tackled this \eg{} by adaptively tracing curved rays~\cite{seyb19nonlinear}.
We do not present any new solutions to this challenge in this work; we sidestep it on a case-by-case basis by point sampling or extracting meshes for visualization only---this is an important problem for future research.

\subsection{Software and Fast Computation}\label{section:ImplementationDetails}
\begin{table}[b]
  \centering
  \caption{Selected timing results. \textit{See supplements for details.} \label{tab:pref_table}}
  \begin{tabular}{l l r r r r}
    \toprule
    \textbf{Name} & \textbf{Figure} & \textbf{Training}  & \textbf{Sim Step}  \\
    Mesh monkey & \autoref{fig:Teaser} & 180 sec  & 51 ms  \\
    SDF key & \autoref{fig:key_modes} & 886 sec  & 107 ms \\
    Gaussian Splat lego & \autoref{fig:Gsplat_results} & 5550 sec & 74 ms \\
    \bottomrule
  \end{tabular}
\end{table}

We implement our method in Pytorch~\cite{paszke2017automatic}, running entirely on the GPU.
Code will be made available upon acceptance to facilitate adoption and clarify details.
All experiments and timings are evaluated on a single RTX 3070. 
Spatial derivatives are evaluated via finite differences, and training gradients are evaluated with standard backpropagation.
For time stepping, we find that assembling Hessians and gradients for Newton steps via automatic differentiation is excessively expensive, and instead assemble them with explicit analytical expressions, albeit still in pure Pytorch code. 
Visualizations are rendered with various packages including Blender, Polyscope, and Houdini as appropriate for each experiment.

\section{Evaluation}
\label{section:Comparison}
\begin{figure}
     \centering
     \vspace{0em}
     \includegraphics{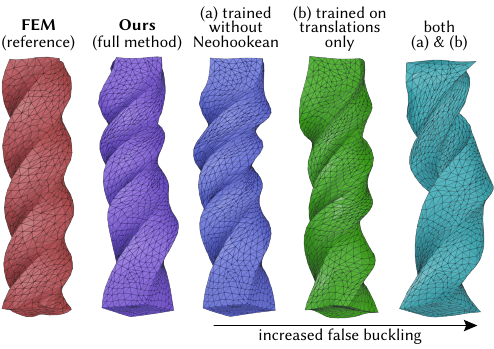}
     \caption{
       An ablation study, on a square bar pinned at the top and twisted at the bottom. Our full method uses the nonlinear Neohookean energy during training, as well as sampling full transformations. Compared to an FEM reference solution (red), removing either component worsens accuracy.
       \label{fig:ablation_t_vs_le}
     }
 \end{figure}

We use standard elastic simulation benchmarks on a prismatic bar to assess the accuracy of our method, ablate its components, and compare to alternatives.
\begin{figure}
    \centering
    \includegraphics{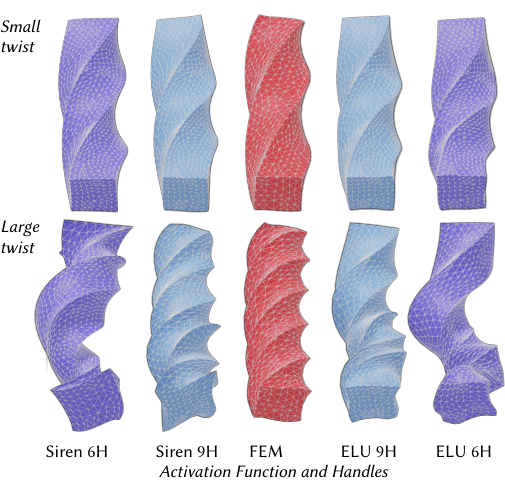}
    \caption{
      \rev{Comparing ELU vs. SIREN \cite{sitzmann2020siren} activation functions on a  large deformation. We experimentally find that using ELU activations, \dl{our simulation satisfies boundary conditions more strictly than when using SIREN.} \label{fig:eluVsSiren}}
    }
\end{figure} 
\paragraph{Validation}
In \autoref{fig:ablation}, we compare a twisted bar simulated with our method to a ground-truth solution from linear tetrahedral FEM. 
Accuracy improves as we increase the number of handles and capacity of the MLP.
\autoref{fig:ablation_t_vs_le} uses the same setup to ablate the use of the Neohookean energy during training, as well as sampling full handle transformations rather than just translations.
Both of these factors introduce increased nonlinear rotation into the training procedure, and as-expected the resulting weights modestly improve accuracy when the bar is in a highly-twisted near-buckling state. 
\autoref{fig:trainingIterationConvergence} shows the training dynamics and loss curve.
Although the loss does not decrease much after 5000 iterations, additional training still visually improves the quality of the simulation. \autoref{fig:eluVsSiren} compares the choice of activation function in the neural network. We find that ELU activation handles boundary conditions better than SIREN (\cite{sitzmann2020siren}).

\begin{figure}
    \centering
    \includegraphics{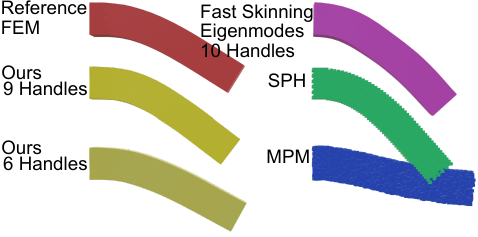}
    \caption{
      \rev{A cantilever bar comparison between a reference linear-tetrahedral, Corotational FEM beam, our method with 6 and 9 handles, Fast Skinning Eigenmodes~\citep{benchekroun2023fast} with 10 handles, SPH~\citep{kugelstadt2021fast} with 5582 particles, and MPM~\citep{hu2019taichi} with 5000 particles and initial grid density of 10. Notice ours matches Fast Skinning Eigenmodes very closely andf exhibits similar numerical coarsening due to reduction when compared to FEM. Simulations are run for 300 steps with timestep $0.01\unit{s}$, with young's modulus $5e6 \unit{Pa}$, poisson ratio $0.45$, density $1000\unit{kg\per m^3}$ using corotational linear elastic material.}\label{fig:manyComparisons}}
\end{figure} 

\paragraph{Comparisons}
In \autoref{fig:manyComparisons}, we show our method generates results that are as comparable to reference FEM as Fast Skinning Eigenmodes (\citet{benchekroun2023fast}) and more than other mesh-free approaches such as SPH (\cite{peer2018implicit}) and MPM\footnote{We tried various particles and grid densities and MPM failed to hold together.} (\cite{Jiang2016mpm}, as implemented in~\cite{hu2019taichi}) with a moderate number of handles. \autoref{fig:Poky_VaryingBeam} likewise compares behavior under sharp contacts.

 \begin{figure}
    \centering
    \vspace{0em}
    \includegraphics{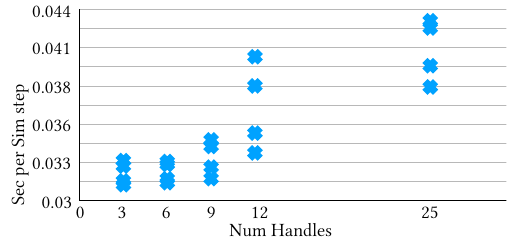}
    \caption{
        \rev{Plotting average time (in ms) per sim step as the number of DOFs increases. Time increases super-linearly due to the dense linear solve in Newton's Method.}
        \label{fig:ablation_dofs_vs_time}
    }
\end{figure}
\begin{wrapfigure}{r}{100pt}
\includegraphics[width=100pt]{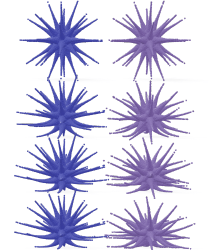}
    \caption{
        \dl{Left: Trained without $\LElastic$ loss. Right: Trained with $\LElastic$.}
    \label{fig:train_with_ortho_only}
    }
    \vspace{-1em}
\end{wrapfigure}

\paragraph{Scaling}
\rev{We observe that training cost depends mainly on the network size rather than the number of handle degrees of freedom. The supplementary table of 144 experiments shows the scaling of training cost as a function of network width and depth.
At simulation time, the dense Newton solves in backward Euler time integration lead to super-linear scaling in the number of handle degrees of freedom (see \autoref{fig:ablation_dofs_vs_time}), though, timesteps generally remain interactive even with many handles.
}
\paragraph{Other Ablations}
\rev{Our training is robust to moderate hyperparameter adjustments in most cases. For example, training on a linear elastic material without an elastic energy scheduler (shown in \autoref{fig:ablation_t_vs_le} (a)) we can achieve visually appealing deformations when simulating a nonlinear material. However, incorporating physics-based energy during training along with the orthogonality term is essential. Training with physics illicits a physics-based response during the simulation as shown in \autoref{fig:train_with_ortho_only}.} 

\rev{Another hyperparameter is the learning rate scheduler, which is optional when using smaller learning rates and more steps, but necessary when starting with a larger learning rate for faster convergence.
We generally avoid tuning hyperparameters per-object in order to show that a common set of parameters is effective in a wide range of settings, with a few exceptions for ablation experiments and particularly large-scale or complex objects.}


\section{Results}

\begin{figure}
     \centering
     \includegraphics[width=\columnwidth]{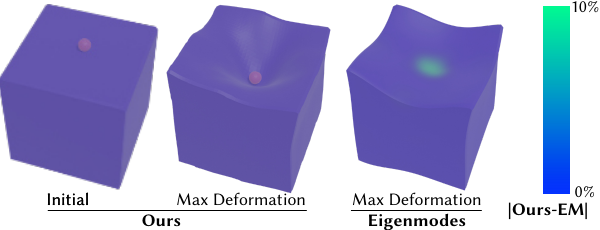}
     \caption{Despite greatly reduced space, our simulations are surprisingly responsive to contacts, which were not seen during the training procedure. For the same number of modes (10), mesh-based eigenmodes fail to capture the sharp nature of the contact. We plot displacement error relative to mesh bounding box diagonal.}
     \label{fig:Poky_VaryingBeam}
\end{figure}

\begin{figure}
     \centering
     \includegraphics[width=\columnwidth]{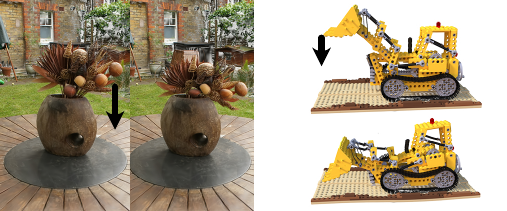}
     \caption{Gaussian splat reconstructions~\citep{kerbl3Dgaussians}, simulated with gravity (left) and free fall of an object (right). \textit{Zoom for details}.}
     \label{fig:Gsplat_results}
\end{figure}
To show the broad applicability of simulation with \ourmethod{}, we demonstrate elastic simulations across a wide variety of representations and data sources.
Please see  the supplemental material for a listing of configurations for all experiments, and two videos giving an overview as well as an extended catalog of results.

\paragraph{Meshes}
To begin, we show simulations on standard triangle and tetrahedral meshes \dl{with occupancy encoded via a signed distance function} (\autoref{fig:Teaser}, \autoref{fig:ablation_t_vs_le}).
These are well-studied in prior simulation methods, though our approach removes assumptions about element quality or mesh cleanness which may be difficult to meet with in-the-wild data, and offers a unified framework applicable to non-mesh data as well.

\begin{figure}
    \centering
    \includegraphics[width=\columnwidth]{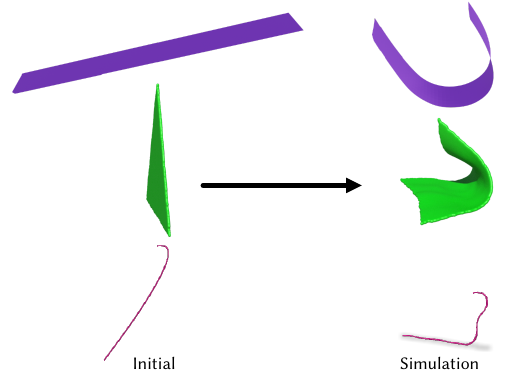}
     \caption{Simulation of thin SDF sheets and strands under gravity and ground collisions.\textit{Please see supplemental video for details.}}
     \label{fig:SDF_Ribbon}
\end{figure}

\paragraph{Signed Distance Functions}
Signed-distance functions \rev{(with occupancy encoded as a scalar field)} are an increasingly popular shape representation, both as artist-constructed analytical functions and learned neural fields. We simulate the entire dataset of SDFs from \citet{Takikawa2022SDF} under gravity and ground contact (\autoref{fig:Teaser}, supplemental video). 
Our adaptive neural field and sampling procedure captures even thin features and codimensional effects (\autoref{fig:SDF_Ribbon}).

\paragraph{Point Clouds}
Our method trivially applies to point clouds \rev{(encoded via ~\citet{Barill2018Winding}'s fast winding numbers)}, interpreting the points as set of samples from the support of $\Occupancy$, as shown in \autoref{fig:PC_Dino_Spike_Tree}. 
The toppling Eagle is a 3D-scanned statue point cloud.

\begin{figure}
     \centering
     \includegraphics[width=\columnwidth]{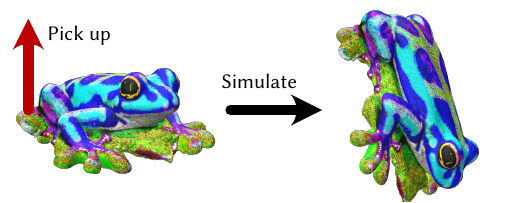}
     \caption{Simulation of a NeRF frog, generated with Magic3D~\cite{lin2023magic3d}. \textit{See the supplemental video for details}.}
     \label{fig:Nerf_Frog}
\end{figure}

\paragraph{NeRF}
Neural radiance fields~\citep{Mildenhall20eccv_nerf} have emerged as a powerful paradigm for reconstruction and machine learning, but working with the resulting content afterwards can be difficult compared to explicit methods.
\ourmethod{} can be applied directly to NeRF representations~\citep{tancik2023nerfstudio} ~\rev{(where occupancy is encoded as a density field)}, even those from recent generative AI systems~\citep{lin2023magic3d} (\autoref{fig:Nerf_Frog}, supplemental video).

\paragraph{Gaussian Splats}
Point splat-based rendering has a long history in computer graphics, and the Gaussian Splat formulation has recently proven to be compelling choice for realtime rendering of reconstructed scenes~\citep{kerbl3Dgaussians}.
Directly simulating Gaussian Splat particles with our method offers fast and high-quality simulations directly on reconstructed data (\autoref{fig:Gsplat_results}).

\begin{figure}
     \centering
     \includegraphics[width=\columnwidth]{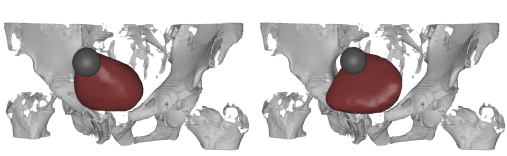}
     \caption{A simulated nonlinear contact on a CT-scanned bladder.}
     \label{fig:CT_Bladder}
\end{figure}

\begin{figure}
     \centering
     \includegraphics[width=\columnwidth]{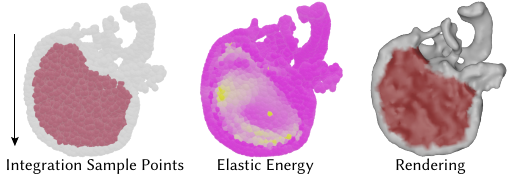}
     \caption{A simulated deformation of a soft brain and stiff skull hitting the floor, directly from CT-scanned geometry.}
     \label{fig:CT_Brain}
\end{figure}

\paragraph{Medical Imaging}
Beyond computer graphics, simulating medical data is deeply important for health applications, but the challenges of real-world scan data mean off-the-shelf simulators are rarely applicable.
We show that \ourmethod{} can be directly applied to meaningful simulations on real CT scan data simply by thresholding the Hounsfield unit for our inside-outside density function $\Occupancy$.
Our results include a bladder from~\cite{BladderCT} with a contact impact (\autoref{fig:CT_Bladder}), and a skull and brain from~\cite{visiblehuman} colliding with a ground plane under gravity (\autoref{fig:CT_Brain}).

 \paragraph{More results}
 We demonstrate our method on a large variety of objects ranging from thin ribbons in \autoref{fig:SDF_Ribbon} to highly intricate Gaussian Splatting and NeRFs scenes in \autoref{fig:Gsplat_results} and \autoref{fig:Nerf_Frog}. Our method can handle large non-linear deformations and deformations from contact as shown in \autoref{fig:Poky_VaryingBeam}. We demonstrate shape-aware simulations on pointclouds in \autoref{fig:PC_Dino_Spike_Tree}. 

\begin{figure}
     \centering
     \includegraphics[width=\columnwidth]{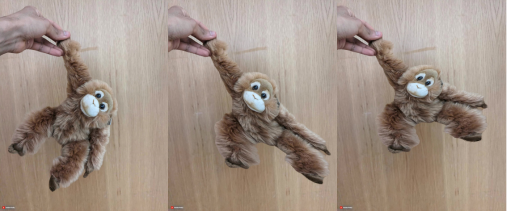}
     \caption{We show preliminary results for simulating elastodynamics in a 2D image. The object is segmented using MIDAS monocular depth~\cite{Ranftl2022} and occluded regions are in-painted using Adobe Firefly.}
     \label{fig:OrangutanSim}
\end{figure}
\section{Conclusion}
We demonstrate \ourmethod{}, a new approach for deformable elastic simulation which is mesh-free, data-free, efficient, and most-importantly agnostic to the underlying 3D representation. 
We have used our method to generate simulations on a breadth of inputs all using the same framework, ranging from CT scans to Gaussian splats to neural implicit objects to triangle meshes, resulting in 140+ simulated outputs in total.

\paragraph{Limitations and  Future Work}

Our neural implicit skinning weight field is fit individually for each object as a pre-processes. Training times are already modest, but could likely be greatly accelerated via grid-based networks which fit similar fields at nearly interactive speeds~\cite{muller2022instant}.

\rev{Although we demonstrate heterogeneous stiffness on several examples (beam~\autoref{fig:ablation_t_vs_le}, skull~\autoref{fig:CT_Brain}, ficus~\autoref{fig:Gsplat_results}), training convergence may be a challenge for objects with complex and highly-variable layered stiffness distributions.} 

In implicit volumetric representations such as NeRFs, rendering from the forward deformation map is nontrivial; future work could address this with invertible deformation representations or deformation-aware rendering schemes~\cite{seyb19nonlinear}. More broadly, the basic \ourmethod{} paradigm could be extended to additional simulated phenomena such as high-frequency secondary effects, fracture, articulated linkages, or extended to other applications in visual computing,  \autoref{fig:OrangutanSim} shows a preliminary result to image editing.

\begin{acks}
This work is funded in part by NSF (1846368, 2313076), NSERC Discovery, Ontario Early Researchers Award Program, the Canada Research Chairs
Program, gifts by Adobe and Autodesk.
We appreciate invaluable feedback from Otman Benchekroun, as well as Abhishek Madan. We thank John Hancock for IT support. Finally, we thank anonymous reviewers for their helpful comments and suggestions.
\end{acks}

\bibliographystyle{ACM-Reference-Format}
\bibliography{references}


\end{document}


\title{Simplicits Supplementary Material}

\author{Vismay Modi}
\affiliation{%
  \institution{University of Toronto}
  \city{Toronto}
  \country{Canada}
}
\email{vismay@cs.toronto.edu}

\author{Nicholas Sharp}
\affiliation{%
  \institution{Nvidia}
  \city{Seattle}
  \country{USA}
}
\email{nmwsharp@gmail.com}

\author{Or Perel}
\affiliation{%
  \institution{Nvidia}
  \city{Tel Aviv}
  \country{Israel}
}
\email{operel@nvidia.com}

\author{Shinjiro Sueda}
\affiliation{%
  \institution{Texas A\&M University}
  \city{College Station}
  \country{USA}
}
\email{sueda@tamu.edu}

\author{David I. W. Levin}
\affiliation{%
  \institution{University of Toronto}
  \city{Toronto}
  \country{Canada}
}
\email{diwlevin@cs.toronto.edu}

\renewcommand{\shortauthors}{Modi et al.}

\maketitle 
\subsection*{Table of Parameters and Times  (in ms) Per Example}
\begin{footnotesize}
\setlength\LTleft{-1in}
\setlength\LTright{-1in}
\begin{longtable}{@{\extracolsep{\fill}}*{13}{|r}|@{}}
    \hline
        \head{1cm}{Input Type} & \head{1cm}{Sim Name} & \head{1cm}{Cubature Pts} & \head{1cm}{Sim Steps} & \head{1cm}{Newton Iters} & \head{1cm}{Log Barrier Its} & \head{1cm}{Per Sim Step (ms)} & \head{1cm}{Handles} & \head{1cm}{Training Steps} & \head{1cm}{LR Start} & \head{1cm}{LR End} & \head{1cm}{Sampling Stdev} & \head{1cm}{Per Train Step (ms)} \\ \hline
        Splats & Garden Splats & 2000 & 100 & 10 & 1 & 77.01 & 40 & 30000 & 0.0010 & 0.0001 & 1.0000 & 18.69 \\ \hline
        Splats & Mic Splats & 2000 & 100 & 10 & 1 & 76.50 & 40 & 30000 & 0.0010 & 0.0001 & 1.0000 & 18.60 \\ \hline
        Splats & Ficus Splats & 2000 & 100 & 10 & 1 & 73.04 & 40 & 30000 & 0.0010 & 0.0001 & 1.0000 & 18.52 \\ \hline
        Splats & Lego Splats & 2000 & 100 & 10 & 1 & 74.59 & 30 & 30000 & 0.0010 & 0.0001 & 1.0000 & 18.62 \\ \hline
        CT & Bladder & 7000 & 150 & 5 & 3 & 1387.27 & 10 & 10000 & 0.0010 & 0.0001 & 1.0000 & 63.02 \\ \hline
        SDF & SkullStrippedBrain & 2000 & 150 & 10 & 1 & 98.27 & 10 & 10000 & 0.0010 & 0.0001 & 1.0000 & 65.48 \\ \hline
        Mesh & 511BeamELU & 1000 & 30 & 10 & 1 & 185.13 & 10 & 10000 & 0.0010 & 0.0001 & 0.1000 & 68.53 \\ \hline
        Nerf & Iron & 2000 & 100 & 10 & 1 & 81.03 & 10 & 10000 & 0.0010 & 0.0001 & 1.0000 & 138.32 \\ \hline
        Nerf & Iron Stiffer & 2000 & 100 & 10 & 1 & 27.76 & 10 & 10000 & 0.0010 & 0.0001 & 1.0000 & 138.32 \\ \hline
        Nerf & Tree & 2000 & 50 & 10 & 1 & 138.70 & 10 & 10000 & 0.0010 & 0.0001 & 1.0000 & 113.90 \\ \hline
        Points & Spike & 2000 & 100 & 10 & 1 & 104.50 & 10 & 10000 & 0.0010 & 0.0001 & 1.0000 & 65.13 \\ \hline
        Points & Tree & 2000 & 50 & 10 & 1 & 136.76 & 10 & 10000 & 0.0010 & 0.0001 & 1.0000 & 63.88 \\ \hline
        Mesh & LargeBox & 5000 & 20 & 10 & 1 & 412.20 & 10 & 10000 & 0.0010 & 0.0001 & 0.1000 & 64.81 \\ \hline
        SDF & Link & 1000 & 100 & 10 & 1 & 72.87 & 10 & 10000 & 0.0010 & 0.0001 & 1.0000 & 151.89 \\ \hline
        SDF & Mandelbulb & 2000 & 50 & 10 & 1 & 138.40 & 10 & 10000 & 0.0010 & 0.0001 & 1.0000 & 65.29 \\ \hline
        SDF & Ribbon & 1000 & 100 & 30 & 1 & 656.64 & 10 & 10000 & 0.0010 & 0.0001 & 1.0000 & 64.07 \\ \hline
        SDF & SimpleSkullBrain & 2000 & 100 & 10 & 1 & 188.36 & 10 & 10000 & 0.0010 & 0.0001 & 1.0000 & 63.63 \\ \hline
        Mesh & Suzanne & 1000 & 50 & 10 & 1 & 180.60 & 5 & 10000 & 0.0010 & 0.0001 & 0.1000 & 51.70 \\ \hline
        SDF & Bezier & 2000 & 70 & 10 & 1 & 119.55 & 10 & 10000 & 0.0010 & 0.0001 & 1.0000 & 77.05 \\ \hline
        SDF & Burger & 2000 & 70 & 10 & 1 & 125.36 & 10 & 10000 & 0.0010 & 0.0001 & 1.0000 & 82.21 \\ \hline
        SDF & Cables & 2000 & 70 & 10 & 1 & 153.78 & 10 & 10000 & 0.0010 & 0.0001 & 1.0000 & 77.44 \\ \hline
        SDF & Capsule & 2000 & 70 & 10 & 1 & 205.43 & 10 & 10000 & 0.0010 & 0.0001 & 1.0000 & 107.84 \\ \hline
        SDF & Castle & 2000 & 70 & 10 & 1 & 194.09 & 10 & 10000 & 0.0010 & 0.0001 & 1.0000 & 65.40 \\ \hline
        SDF & Chain & 2000 & 70 & 10 & 1 & 133.15 & 10 & 10000 & 0.0010 & 0.0001 & 1.0000 & 96.51 \\ \hline
        SDF & Cheese & 2000 & 70 & 10 & 1 & 215.03 & 10 & 10000 & 0.0010 & 0.0001 & 1.0000 & 87.56 \\ \hline
        SDF & Cone & 2000 & 70 & 10 & 1 & 218.63 & 10 & 10000 & 0.0010 & 0.0001 & 1.0000 & 88.57 \\ \hline
        SDF & Cube & 2000 & 70 & 10 & 1 & 194.06 & 10 & 10000 & 0.0010 & 0.0001 & 1.0000 & 88.92 \\ \hline
        SDF & Cybertruck & 2000 & 70 & 10 & 1 & 196.28 & 10 & 10000 & 0.0010 & 0.0001 & 1.0000 & 97.35 \\ \hline
        SDF & Cylinder & 2000 & 70 & 10 & 1 & 185.83 & 10 & 10000 & 0.0010 & 0.0001 & 1.0000 & 81.67 \\ \hline
        SDF & Dalek & 2000 & 70 & 10 & 1 & 200.16 & 10 & 10000 & 0.0010 & 0.0001 & 1.0000 & 61.04 \\ \hline
        SDF & Dinosaur & 2000 & 70 & 10 & 1 & 126.38 & 10 & 10000 & 0.0010 & 0.0001 & 1.0000 & 79.02 \\ \hline
        SDF & Dodecahedron & 2000 & 70 & 10 & 1 & 138.51 & 10 & 10000 & 0.0010 & 0.0001 & 1.0000 & 141.56 \\ \hline
        SDF & Elephant & 2000 & 70 & 10 & 1 & 110.37 & 10 & 10000 & 0.0010 & 0.0001 & 1.0000 & 100.30 \\ \hline
        SDF & Fish & 2000 & 70 & 10 & 1 & 122.52 & 10 & 10000 & 0.0010 & 0.0001 & 1.0000 & 112.89 \\ \hline
        SDF & Gear & 2000 & 70 & 10 & 1 & 116.52 & 10 & 10000 & 0.0010 & 0.0001 & 1.0000 & 82.61 \\ \hline
        SDF & Girl & 2000 & 70 & 10 & 1 & 119.23 & 10 & 10000 & 0.0010 & 0.0001 & 1.0000 & 60.56 \\ \hline
        SDF & GrandPiano & 2000 & 70 & 10 & 1 & 104.82 & 10 & 10000 & 0.0010 & 0.0001 & 1.0000 & 75.75 \\ \hline
        SDF & Helix & 2000 & 70 & 10 & 1 & 110.33 & 10 & 10000 & 0.0010 & 0.0001 & 1.0000 & 87.81 \\ \hline
        SDF & Hexprism & 2000 & 70 & 10 & 1 & 112.23 & 10 & 10000 & 0.0010 & 0.0001 & 1.0000 & 87.65 \\ \hline
        SDF & HumanHead & 2000 & 70 & 10 & 1 & 107.05 & 10 & 10000 & 0.0010 & 0.0001 & 1.0000 & 87.93 \\ \hline
        SDF & HumanSkull & 2000 & 70 & 10 & 1 & 118.00 & 10 & 10000 & 0.0010 & 0.0001 & 1.0000 & 87.78 \\ \hline
        SDF & Icosahedron & 2000 & 70 & 10 & 1 & 114.67 & 10 & 10000 & 0.0010 & 0.0001 & 1.0000 & 87.77 \\ \hline
        SDF & Jellyfish & 2000 & 70 & 10 & 1 & 108.75 & 10 & 10000 & 0.0010 & 0.0001 & 1.0000 & 87.88 \\ \hline
        SDF & Jetfighter & 2000 & 70 & 10 & 1 & 116.12 & 10 & 10000 & 0.0010 & 0.0001 & 1.0000 & 87.90 \\ \hline
        SDF & Julia & 2000 & 70 & 10 & 1 & 112.98 & 10 & 10000 & 0.0010 & 0.0001 & 1.0000 & 88.20 \\ \hline
        SDF & Key & 2000 & 70 & 10 & 1 & 107.79 & 10 & 10000 & 0.0010 & 0.0001 & 1.0000 & 88.68 \\ \hline
        SDF & Knob & 2000 & 70 & 10 & 1 & 115.73 & 10 & 10000 & 0.0010 & 0.0001 & 1.0000 & 88.12 \\ \hline
        SDF & Lamborghini & 2000 & 70 & 10 & 1 & 118.07 & 10 & 10000 & 0.0010 & 0.0001 & 1.0000 & 87.94 \\ \hline
        SDF & Mandelbulb & 2000 & 70 & 10 & 1 & 112.28 & 10 & 10000 & 0.0010 & 0.0001 & 1.0000 & 88.21 \\ \hline
        SDF & MantaRay & 2000 & 70 & 10 & 1 & 112.69 & 10 & 10000 & 0.0010 & 0.0001 & 1.0000 & 94.02 \\ \hline
        SDF & Mech & 2000 & 70 & 10 & 1 & 115.23 & 10 & 10000 & 0.0010 & 0.0001 & 1.0000 & 88.05 \\ \hline
        SDF & Menger & 2000 & 70 & 10 & 1 & 125.33 & 10 & 10000 & 0.0010 & 0.0001 & 1.0000 & 88.52 \\ \hline
        SDF & Mobius & 2000 & 70 & 10 & 1 & 112.79 & 10 & 10000 & 0.0010 & 0.0001 & 1.0000 & 87.86 \\ \hline
        SDF & Mountain & 2000 & 70 & 10 & 1 & 121.30 & 10 & 10000 & 0.0010 & 0.0001 & 1.0000 & 88.34 \\ \hline
        SDF & Mushroom & 2000 & 70 & 10 & 1 & 111.75 & 10 & 10000 & 0.0010 & 0.0001 & 1.0000 & 89.19 \\ \hline
        SDF & Octabound & 2000 & 70 & 10 & 1 & 97.48 & 10 & 10000 & 0.0010 & 0.0001 & 1.0000 & 88.44 \\ \hline
        SDF & Octahedron & 2000 & 70 & 10 & 1 & 103.35 & 10 & 10000 & 0.0010 & 0.0001 & 1.0000 & 87.93 \\ \hline
        SDF & Oldcar & 2000 & 70 & 10 & 1 & 118.36 & 10 & 10000 & 0.0010 & 0.0001 & 1.0000 & 88.18 \\ \hline
        SDF & PixarMike & 2000 & 70 & 10 & 1 & 105.82 & 10 & 10000 & 0.0010 & 0.0001 & 1.0000 & 87.93 \\ \hline
        SDF & Pyramid & 2000 & 70 & 10 & 1 & 107.72 & 10 & 10000 & 0.0010 & 0.0001 & 1.0000 & 87.83 \\ \hline
        SDF & Rock & 2000 & 70 & 10 & 1 & 110.81 & 10 & 10000 & 0.0010 & 0.0001 & 1.0000 & 88.10 \\ \hline
        SDF & Rooks & 2000 & 70 & 10 & 1 & 114.72 & 10 & 10000 & 0.0010 & 0.0001 & 1.0000 & 87.72 \\ \hline
        SDF & Roundbox & 2000 & 70 & 10 & 1 & 114.37 & 10 & 10000 & 0.0010 & 0.0001 & 1.0000 & 88.08 \\ \hline
        SDF & Serpinski & 2000 & 70 & 10 & 1 & 105.84 & 10 & 10000 & 0.0010 & 0.0001 & 1.0000 & 87.83 \\ \hline
        SDF & Snail & 2000 & 70 & 10 & 1 & 117.75 & 10 & 10000 & 0.0010 & 0.0001 & 1.0000 & 87.80 \\ \hline
        SDF & Snake & 2000 & 70 & 10 & 1 & 113.79 & 10 & 10000 & 0.0010 & 0.0001 & 1.0000 & 87.87 \\ \hline
        SDF & Sphere & 2000 & 70 & 10 & 1 & 104.47 & 10 & 10000 & 0.0010 & 0.0001 & 1.0000 & 87.93 \\ \hline
        SDF & Spike & 2000 & 70 & 10 & 1 & 115.03 & 10 & 10000 & 0.0010 & 0.0001 & 1.0000 & 87.79 \\ \hline
        SDF & Tardigrade & 2000 & 70 & 10 & 1 & 112.30 & 10 & 10000 & 0.0010 & 0.0001 & 1.0000 & 87.99 \\ \hline
        SDF & Teapot & 2000 & 70 & 10 & 1 & 110.79 & 10 & 10000 & 0.0010 & 0.0001 & 1.0000 & 92.23 \\ \hline
        SDF & Temple & 2000 & 70 & 10 & 1 & 118.80 & 10 & 10000 & 0.0010 & 0.0001 & 1.0000 & 87.69 \\ \hline
        SDF & Tetrahedron & 2000 & 70 & 10 & 1 & 120.33 & 10 & 10000 & 0.0010 & 0.0001 & 1.0000 & 87.99 \\ \hline
        SDF & TieFighter & 2000 & 70 & 10 & 1 & 108.20 & 10 & 10000 & 0.0010 & 0.0001 & 1.0000 & 87.92 \\ \hline
        SDF & Torus & 2000 & 70 & 10 & 1 & 105.98 & 10 & 10000 & 0.0010 & 0.0001 & 1.0000 & 87.93 \\ \hline
        SDF & Tree & 2000 & 70 & 10 & 1 & 105.15 & 10 & 10000 & 0.0010 & 0.0001 & 1.0000 & 87.77 \\ \hline
        SDF & Trefoil & 2000 & 70 & 10 & 1 & 107.46 & 10 & 10000 & 0.0010 & 0.0001 & 1.0000 & 87.72 \\ 
        \label{table:all_experiments}
\end{longtable}
\end{footnotesize}